\documentclass[conference]{IEEEtran}
\IEEEoverridecommandlockouts
\usepackage{cite}
\usepackage{amsmath,amssymb,amsfonts}
\usepackage{algorithmic}
\usepackage{graphicx}
\usepackage{textcomp}
\usepackage{xcolor}
\usepackage{hyperref}
\hypersetup{
    colorlinks=true,
    linkcolor=black,
    filecolor=black,      
    urlcolor=black,
}

\def\BibTeX{{\rm B\kern-.05em{\sc i\kern-.025em b}\kern-.08em
    T\kern-.1667em\lower.7ex\hbox{E}\kern-.125emX}}
\begin{document}

\title{Evolutionary Swarm Robotics: Dynamic Subgoal-Based Path Formation and Task Allocation for Exploration and Navigation in Unknown Environments\\
}
\author{
\IEEEauthorblockN{Lavanya Ratnabala\textsuperscript{1}, Robinroy Peter\textsuperscript{2}, E.Y.A. Charles\textsuperscript{3}}
\IEEEauthorblockA{
\textsuperscript{1}\textit{Department of Computer Science, University Of Jaffna}\\
Jaffna, Sri Lanka\\
lavanyaratnabala@gmail.com}
\IEEEauthorblockA{
\textsuperscript{2}\textit{Department of Computer Science, University Of Jaffna}\\
Jaffna, Sri Lanka \\
robinroy.peter@gmail.com}
\IEEEauthorblockA{
\textsuperscript{3}\textit{Department of Computer Science, University Of Jaffna}\\
Jaffna, Sri Lanka\\
charles.ey@univ.jfn.ac.lk}

}

\maketitle

\begin{abstract}
This research paper addresses the challenges of exploration and navigation in unknown environments from an evolutionary swarm robotics perspective. Path formation plays a crucial role in enabling cooperative swarm robots to accomplish these tasks. The paper presents a method called the sub-goal-based path formation, which establishes a path between two different locations by exploiting visually connected sub-goals. Simulation experiments conducted in the Argos simulator demonstrate the successful formation of paths in the majority of trials.

Furthermore, the paper tackles the problem of inter-collision (traffic) among a large number of robots engaged in path formation, which negatively impacts the performance of the sub-goal-based method. To mitigate this issue, a task allocation strategy is proposed, leveraging local communication protocols and light signal-based communication. The strategy evaluates the distance between points and determines the required number of robots for the path formation task, reducing unwanted exploration and traffic congestion. The performance of the sub-goal-based path formation and task allocation strategy is evaluated by comparing path length, time, and resource reduction against the A* algorithm. The simulation experiments demonstrate promising results, showcasing the scalability, robustness, and fault tolerance characteristics of the proposed approach.
\end{abstract}


\section{Introduction}
\begin{figure}
    \centering
    \includegraphics[width=1\columnwidth]{ 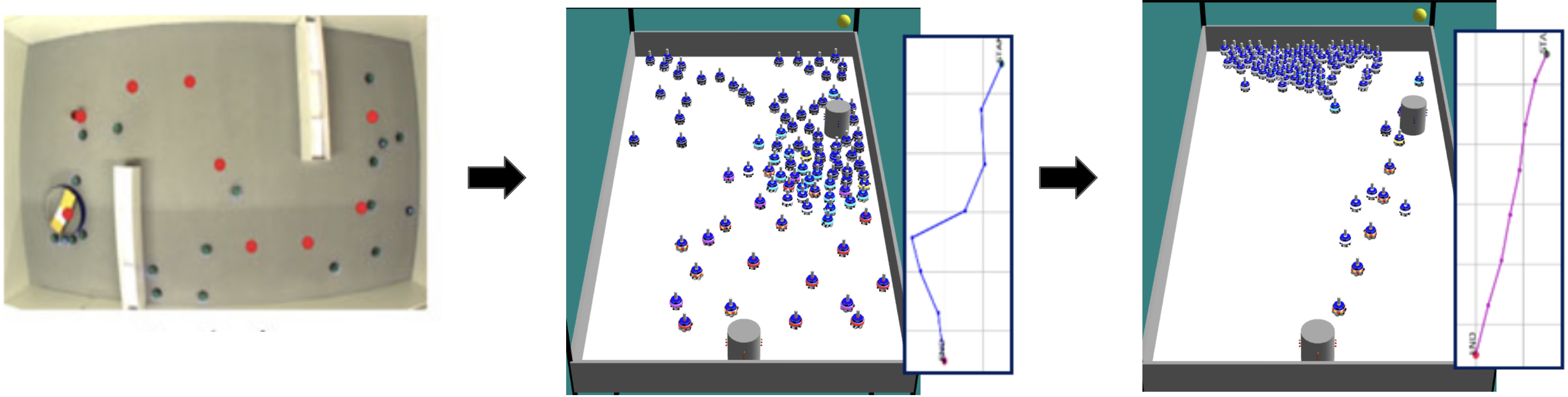}
    \caption{Path Formation Flow}
    \label{fig:protocol}
\end{figure}
\footnote{\href{https://github.com/Robinroy-peter/Dynamic-Subgoal-Based-Path-Formation-and-Task-Allocation-Exploration-Navigation-Unknown-Environments.git}{GitHub Repository: Dynamic Subgoal-Based Path Formation}}

Robotics has emerged as a captivating field of research, experiencing continuous growth over the past few decades. The development of highly sophisticated robots capable of handling the demanding challenges of the real world efficiently is no easy feat. Instead of relying on a single advanced robot, the use of multiple robots becomes necessary to tackle vast and complex tasks effectively. These systems, known as multi-robot systems (MRS), involve the coordination and cooperation of multiple entities, primarily robots, working together to achieve common goals. In some cases, human beings and centralized systems may also be part of these systems. However, effectively coordinating different agents in an MRS poses several challenges, particularly in terms of autonomy and human factors. The deployment and operation of these systems in real-world scenarios require a broad sense of autonomy, wherein robots possess enhanced capabilities and intelligence to operate under adverse conditions for extended periods.
\begin{figure}
    \centering
    \includegraphics[width=0.5\columnwidth]{ 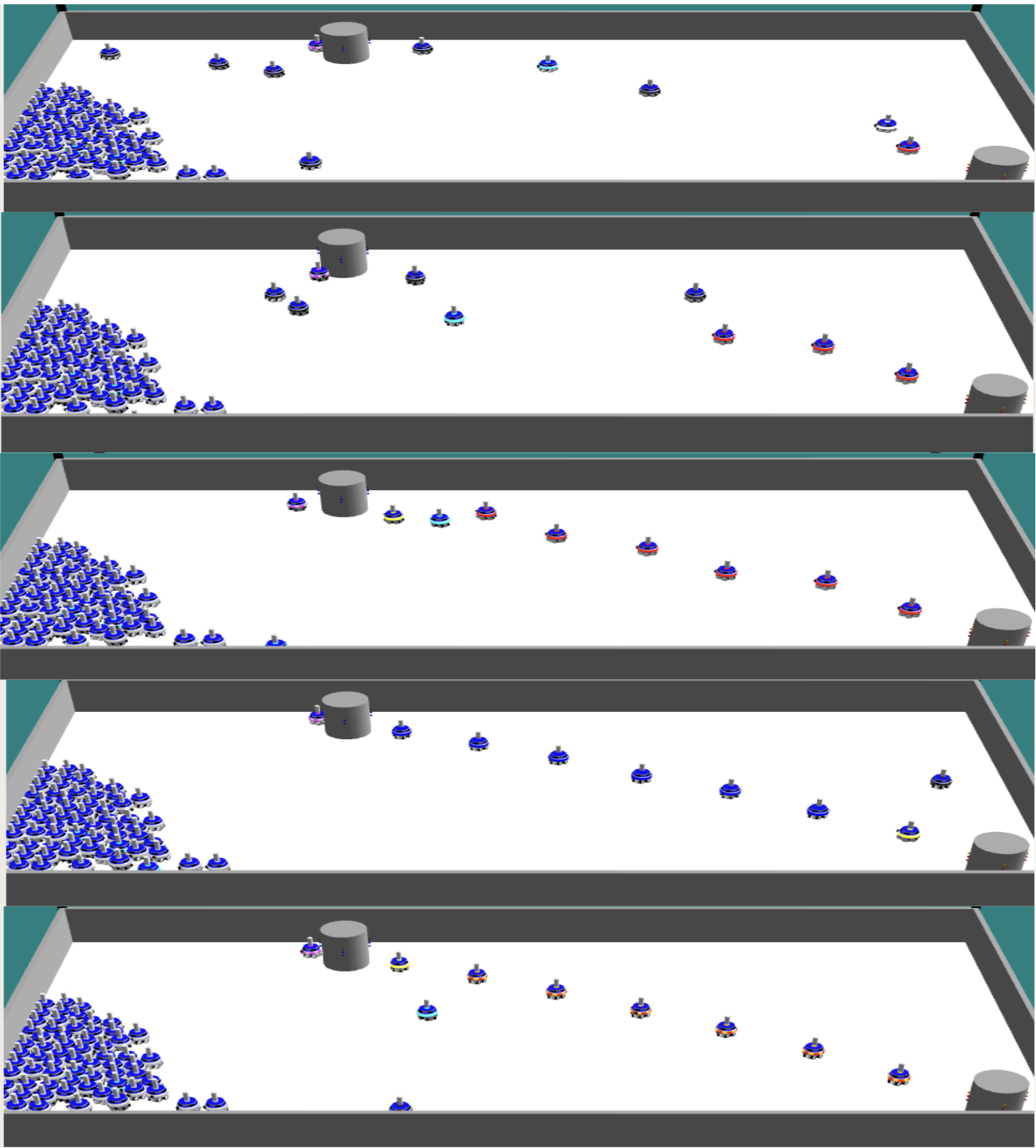}
    \caption{Path Formation}
    \label{fig:protocol}
\end{figure}
Within the realm of MRS, swarm robotics focuses on the utilization of a large number of simple robots. This branch of robotics draws inspiration from swarm intelligence observed in social animals such as ants, bees, and fishes. These creatures demonstrate remarkable abilities to handle difficult tasks collectively, surpassing what individuals can achieve alone. Swarm robotics embraces this concept as its foundation, where global behavior emerges from the coordination of simple individual behaviors or rules, akin to self-organization in natural swarm systems. Swarm robots possess limited sensing and actuating capabilities, with communication primarily limited to local interactions, such as those with neighboring robots and the environment. In scenarios requiring extensive area coverage, a multitude of robots, ranging from hundreds to thousands or more, can be deployed. In such cases, centralized organization can lead to system failures due to information overflow. However, swarm robotic systems do not rely on a centralized agent or controller, as macroscopic or global behavior arises from decentralized, local interactions. Swarm robotic systems exhibit scalability, ensuring their functioning and efficiency remain unaffected by changes in the number of robots. With a foundation in local sensing, swarm systems are adaptable and flexible, able to respond to disturbances within the working environment. The use of simple robots in swarm robotics leads to reduced production costs compared to the development of a single complex robot, thanks to their small size, simple shape, and computational complexity. Key factors in swarm robot design are their miniature form factor and cost-effectiveness. Each member of a swarm team must be resource-efficient and energy-conscious.

Efficient cooperative operation in a swarm robot system necessitates the use of a novel method for cooperative subgoal-based path formation using swarms of robots. This method should enable obstacle avoidance during the cooperative subgoal-based path formation, even in the absence or with limited inter-robot communication. However, employing a large number of robots can lead to decreased performance due to inter-robot collisions and traffic congestion. Task allocation methods can enhance efficiency by forming the shortest and quickest paths. The objective is to assign tasks to the robots in a manner that optimizes cooperation and achieves the global objective more efficiently. In our case, the effective assignment of robots for the path formation task is crucial, and we propose to employ task allocation techniques. In the context of this paper, the two tasks at hand are resting and path formation.
\section{problem Description}
In the realm of cooperative motions and applications involving a group of robots, the generation of a navigable path presents a crucial and complex challenge. Existing approaches for generating paths between two unknown targets have not been specifically designed for other cooperative tasks, such as cooperative navigation or pushing. Furthermore, the quality and applicability of these approaches in other swarms of robotic cooperative tasks have not been thoroughly addressed. Since complex tasks with swarms of robots often require a combination of various cooperative tasks, it becomes essential to have a cooperative path generation algorithm that can generate paths applicable to multiple cooperative robot swarm tasks.

However, a significant issue arises when a large number of robots work together to form a path within a confined area: traffic congestion. The presence of traffic congestion can severely impact the effectiveness and efficiency of the overall system. As robots move randomly with limited visual and communication range, they may become lost or venture into unwanted areas, hindering the path formation process.

To overcome these challenges, the problem at hand calls for the utilization of task allocation techniques to ensure that only the necessary robots are involved in the path formation task. By dynamically allocating tasks to specific robots, the system can prevent unnecessary exploration and minimize the occurrence of traffic congestion.
\section{Related works}
Collective navigation involves a robot reaching a destination by traversing an unknown environment with the assistance of other robots. This task is typically accomplished using communication techniques and finite-state machines. In one study, researchers proposed a strategy for transporting a large object to a goal using a substantial number of mobile robots, which are considerably smaller than the object itself. The robots only push the object at positions where the direct line of sight to the goal is obstructed by the object \cite{b1}. In their future work, they discussed the use of sub-goal-based path formation to push the object through complex environments with numerous obstacles and larger scales. The proposed method involved setting intermediate goals between the starting point and the goal point. Robots would follow the path discovered and push the object toward each sub-goal consecutively until reaching the final goal.

Inspired by natural phenomena, such as ant foraging, researchers have proposed several methods for creating efficient paths between unknown targets \cite{b22}, utilizing artificial pheromones. These methods employ various techniques to generate artificial pheromones, including the release of alcohol, heat, odor, visual marks, or RFID tags. While these methods have shown efficacy in creating efficient paths, artificial pheromone systems may not be reliable in more realistic scenarios. As an alternative, a novel approach involving local IR (infrared) range bearing has been proposed \cite{b23}.

Another efficient approach to path formation involves creating a chain using physically connected or non-connected field-based methods. Two different controllers, vector-field and chain, have been proposed to form paths. The process begins with robots starting to form a path from the prey once it has been detected. The direction of the vector field depends on the LED light direction of the robots already part of the path. Each robot probabilistically decides whether to join the path or not. Additionally, an evolutionary-based approach has been suggested for generating paths between two targets, as evolutionary robotics has shown promising results in solving cooperative tasks in swarm robotics.

Various strategies exist to address task allocation in swarm robots. In a multi-foraging scenario, researchers proposed a task allocation model using the distributed bees algorithm (DBA) \cite{b10}, inspired by the foraging behavior of natural bee colonies. Robots were designed to use broadcast communication to inform other robots within range about the estimated location and quality of discovered targets in a decentralized manner. Another method \cite{b11} eliminates the need for global knowledge, communication between robots, and centralized components, relying solely on local interactions and individual robot perceptions. In another approach \cite{b12}, each robot has a probability of employing task partitioning, defined by sigmoid functions. Two novel-based approaches have been proposed to assign robots to announced tasks \cite{b13}. One approach relies on simple reactive mechanisms based on light signal interactions, while the other employs a more advanced gossip-based communication scheme to announce task requirements among the robots. A self-organizing method has also been proposed for allocating a swarm of robots to perform a foraging task with sequentially dependent sub-tasks \cite{b18}, based on the response threshold model. In this method, each robot updates its response threshold based on the task demand and the number of neighboring robots performing the task.

The Argos simulator \cite{b16} is a notable multi-robot simulator capable of simulating up to 10,000 robots simultaneously. One key feature of the Argos simulator is its ability to apply different physics engines to different regions of the arena and run them in parallel. Argos has been specifically designed and developed for swarm robotics research.
\section{methodology}
In this study, we propose a sub-goal-based path formation method using Finite State Machines (FSMs). Building upon previous approaches that utilize robotic chains, our method introduces dynamic robots within a sub-goal without requiring local intercommunication among them. We adapt foraging concepts to form a path between a nest and a goal, allowing for increased flexibility in various environments. Unlike previous methods, our approach addresses path formation using swarm robots in obstacle-ridden and complex environments. To improve path efficiency, we introduce Recovery Behavior/Hidden Location Identifier robots that indicate and complete the path effectively. Two types of alignment processes are employed: one from the starting point to the goal, and another from the goal back to the starting position.
\begin{figure}
    \centering
    \includegraphics[width=0.07\columnwidth]{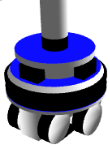}
    \includegraphics[width=0.13\columnwidth]{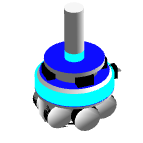}
    \includegraphics[width=0.07\columnwidth]{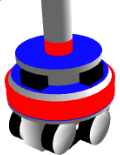}
     \includegraphics[width=0.07\columnwidth]{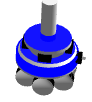}
    \includegraphics[width=0.07\columnwidth]{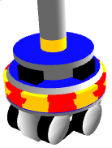}
       \includegraphics[width=0.08\columnwidth]{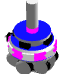}
        \includegraphics[width=0.12\columnwidth]{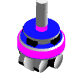}
         \includegraphics[width=0.07\columnwidth]{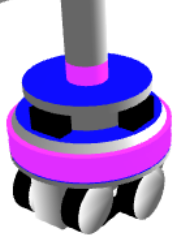}
          \includegraphics[width=0.12\columnwidth]{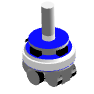}
    \caption{robots color indication in exploring, return to nest, subgoal, 1st optimization, 2nd optimization,goal founder, recovery, decision maker and resting}
    \label{fig:colors}
\end{figure}

\begin{figure}
    \centering
    \includegraphics[width=1.2\columnwidth]{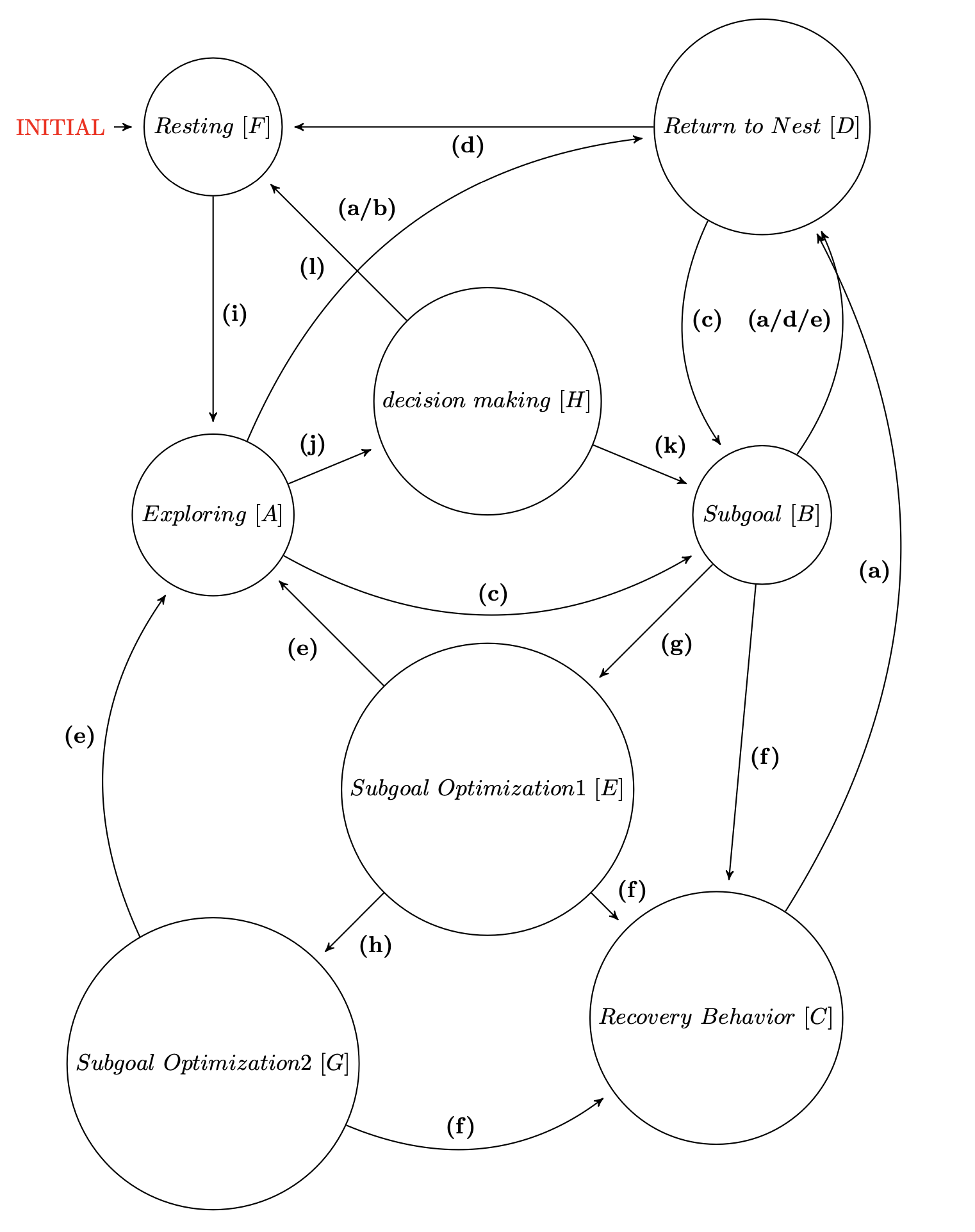}
    \caption{State model of path formation}
    \label{fig:statemodel1}
\end{figure}

\begin{table}[h!]
\caption{States description}
    \renewcommand{\arraystretch}{1.5}
	\begin{tabular}{|p{0.5in}|p{2.777in}|} 
	\hline
	States&Description\\
	\hline
		Resting&robots taking rest in place for predefined steps \\
		Exploring& Searching for the goal from the starting point. Searching will happen opposite potential fields from the starting point. Robots have max step size if robots fail to find the goal then roots change the state to returning to nest through this robot will reach the goal using potential field \\
		Subgoal&When robot find a goal or subgoal its try to form as a subgoal in between starting point and detected subgoal or goal \\
		Return To Nest& Searching for the goal from the starting point(nest).searching will happen opposite potential field from the starting point.Robots have max step sizes if robots fail to find the goal point then the robots change the state to returning to nest \\
        Decision Making&When a robot find a goal or robots find goal finding robot go to this state. Here robots decide whether to go path formation or resting based on local communication protocols.\\
		Recovery Behavior& In subgoal forming state when robot lost the visibility of detected subgoal or goal in the visible range then the robot change as a recovery behavior robot to help other robots not to enter that black spot around some radius \\
		Subgoal Optimization1&When the path formation gets complete at the point of start, then the first subgoal robot which is near the starting point starts to align itself to form in a straight line in between starting point \\
		subgoal Optimization2&When the optimization 1 is completed in goal then 2nd optimization will start from the robot which is near to goal, it's like optimization1, but it will happen from goal to start\\
    \hline
	\end{tabular}
	\label{tab:State Table}
\end{table}

\begin{table}[h!]
\caption{State Transitions}
    \centering
    \begin{tabular}{|r|p{2in}|}
	\hline
	Transitions&Description\\ 
	\hline
		a&Path formation successfully completed\\
		b& Un successful exploration \\
		c& Found the goal/ subgoal\\
		d&Reach nest\\
		e& Found another swarm robot in B,E,G state to archive the end position of that state\\
		f&Lost the visibility of goal/ subgoal in that visible range\\
		g&Successfully form as a subgoal and detect the colors pattern to move state E\\
		h& Successfully finished the Optimization1 and detect the colors pattern to move state G\\
		i&Resting time period get finished\\
  
		j&Once a agent find a goal its goes to this state and ask other agents to join this in order to do the agents allocation  for the path formation tasks \\
		k& Found target and return to nest for making decision about how many robots need to go for path formation for that minimum exploring time\\
	\hline
	\end{tabular}
	\label{tab:Transitions}
\end{table}
A state model of the path formation process is depicted in Figure \ref{fig:statemodel1}. The model consists of various states, each representing a specific behavior. Table 1 provides descriptions of the different states, including Resting, Exploring, Subgoal, Return to Nest, Recovery Behavior, Subgoal Optimization 1, and Subgoal Optimization 2.

\subsection{Experimental setup}
The experimental setup for this study revolves around the s-bot, which has been developed as part of the SWARM-BOTS project. While an individual s-bot may have limited capabilities, a swarm of s-bots is designed to overcome these limitations and operate efficiently. Since the physical s-bots are still under construction, all experiments were conducted using simulation.

Prototype s-bots were constructed, and their specifications were used to develop the simulation software Argos. This software provides a 2D/3D simulation environment that takes into account the dynamics and collisions of rigid bodies. By utilizing Argos, we were able to simulate the behavior of the s-bots and test our sub-goal-based path formation method effectively.
\begin{figure}[h!]
    \centering
    \includegraphics[width=0.4\columnwidth]{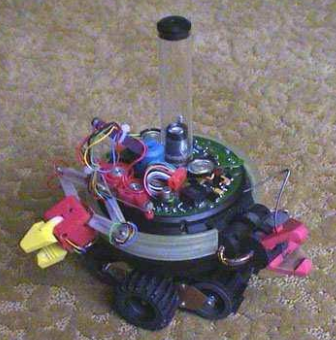}
    \caption{s-bot}
    \label{sf:prototype}
\end{figure}
Table 2 presents the state transitions within the model, providing an explanation of the conditions and events that trigger a transition from one state to another. Each transition is assigned a label (a to i) and corresponds to specific situations such as the successful completion of path formation, unsuccessful exploration, discovery of a goal or sub-goal, return to the nest, encounter with another swarm robot in certain states, loss of visibility of the goal or sub-goal, successful formation of a sub-goal, completion of the first optimization, and completion of the resting time period.

The finite state machine model depicted in Figure \ref{fig:statemodel1}, along with the detailed information provided in Tables 1 and 2, serves as the foundation for our sub-goal-based path formation method.
\subsection{Subgoal Formation}
In the subgoal formation phase, the robots begin by exploring the environment in order to find a goal. During the exploration state, each robot engages in random exploration, increasing its distance from the nest. If the minimum unsuccessful exploration time is exceeded without finding the goal, the robot returns to the starting position to initiate the next exploration.

\begin{figure}
    \centering
    \includegraphics[width=0.5\columnwidth]{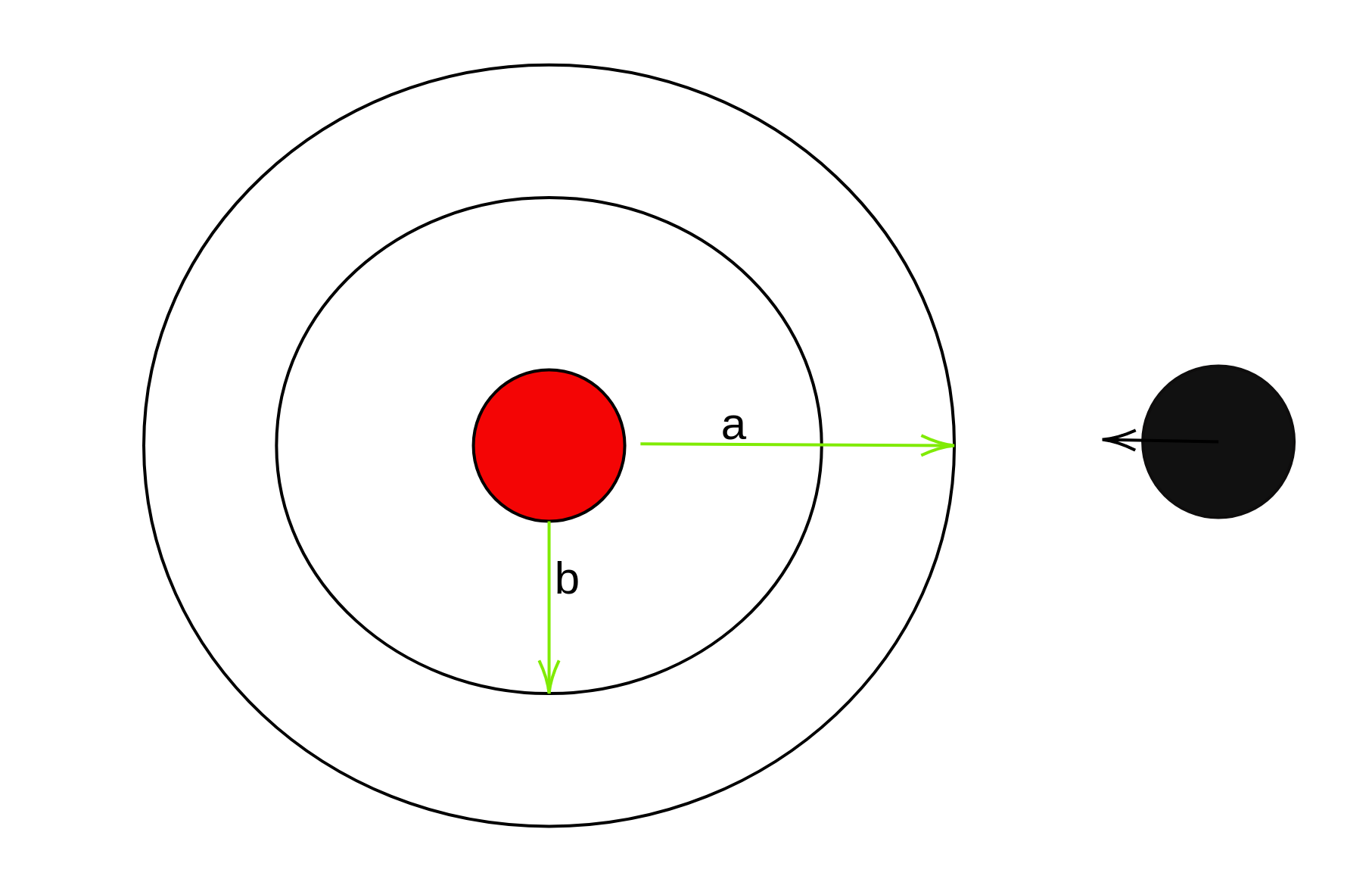}
    \caption{Descriptions of explorer robot parameter}
    \label{fig:parameter1}
\end{figure}

Once a robot detects the goal within a range of 30 in Argos, it transitions to the subgoal state-changing range. At this point, the robot emits a color signal, specifically white, and moves towards the starting point using a potential field approach. It positions itself as a static subgoal at a range of 70 in Argos. This process is illustrated in Figure \ref{fig:parameter1}.

\begin{figure}
    \centering
    \includegraphics[width=0.3\columnwidth]{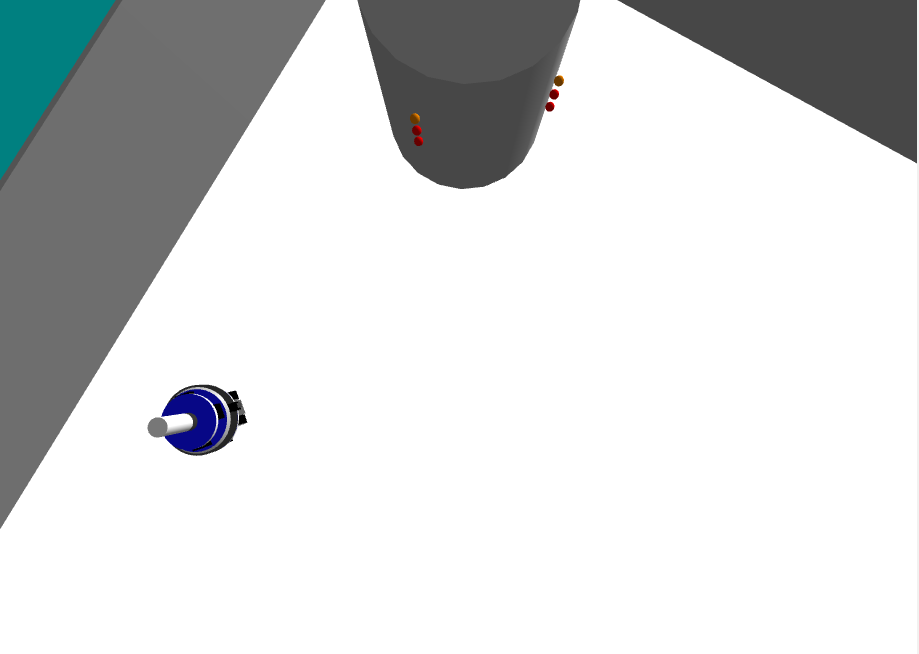}
    \includegraphics[width=0.3\columnwidth]{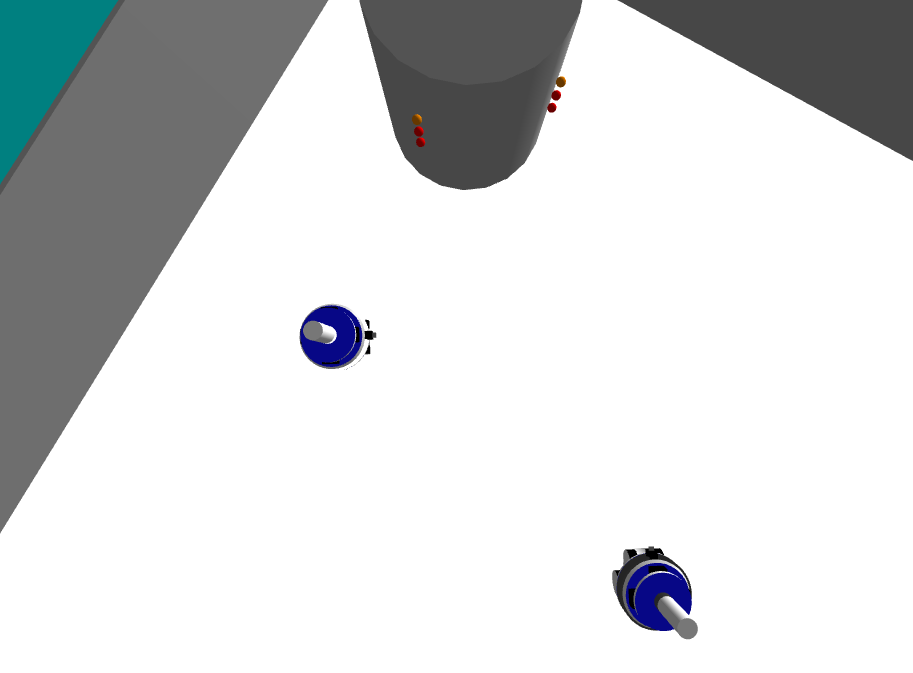}
    \includegraphics[width=0.3\columnwidth]{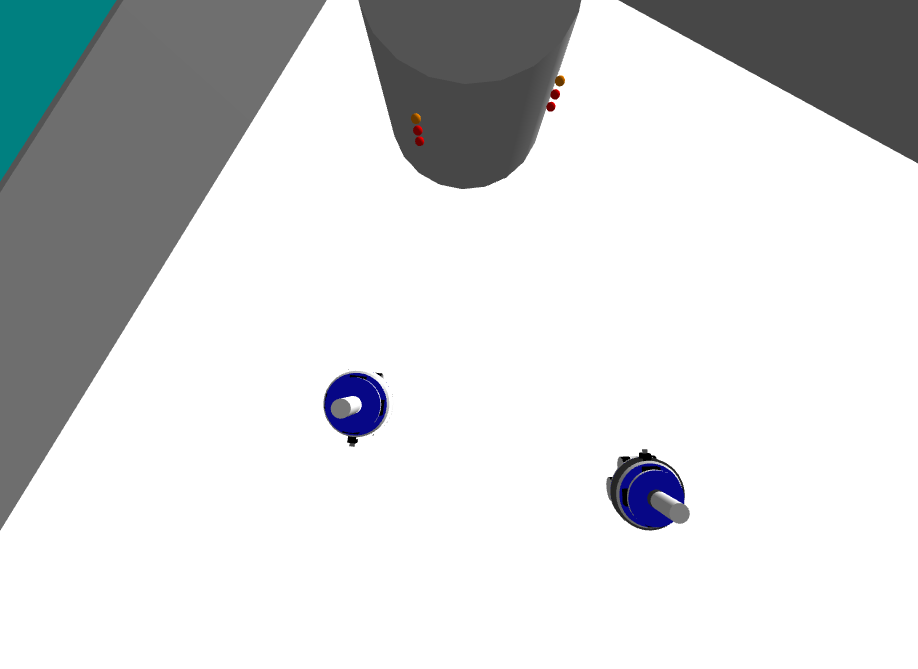}
    \includegraphics[width=0.3\columnwidth]{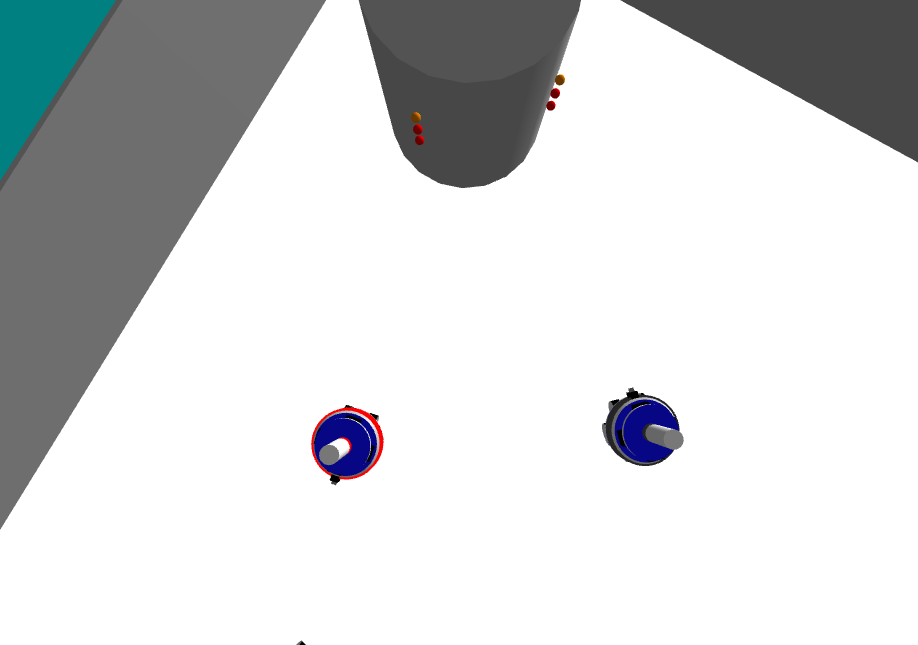}
    \caption{Subgoal formation process from goal}
    \label{fig:subgoalformation1}
\end{figure}

\begin{figure}
    \centering
    \includegraphics[width=0.25\columnwidth]{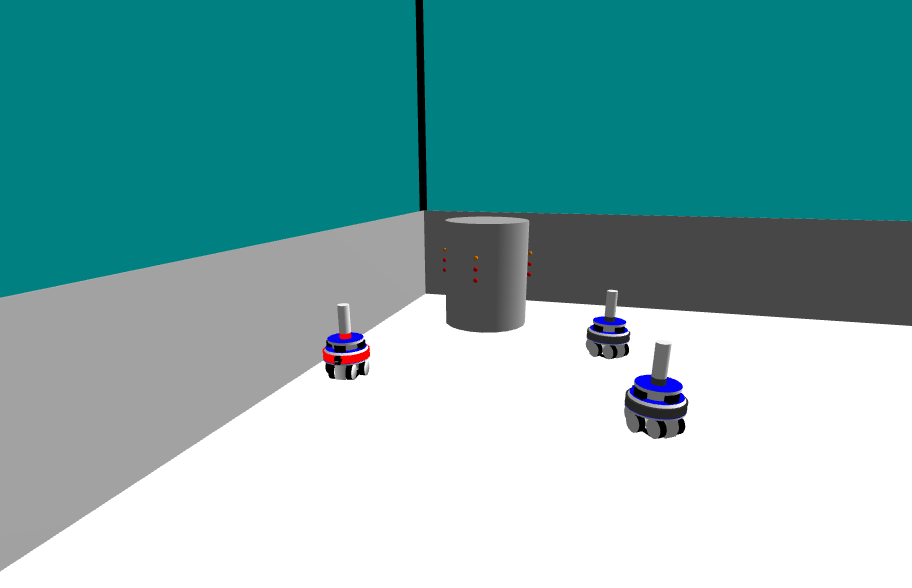}
    \includegraphics[width=0.25\columnwidth]{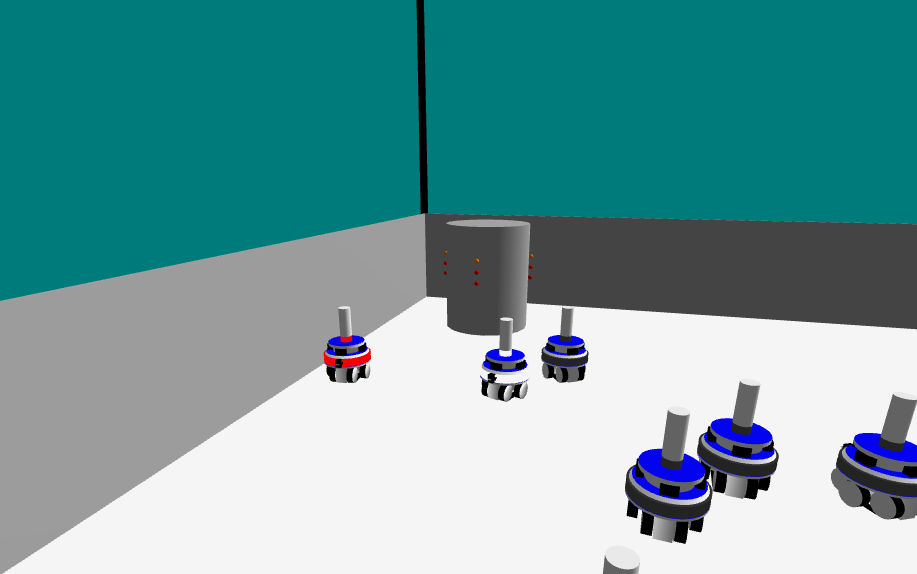}
    \includegraphics[width=0.25\columnwidth]{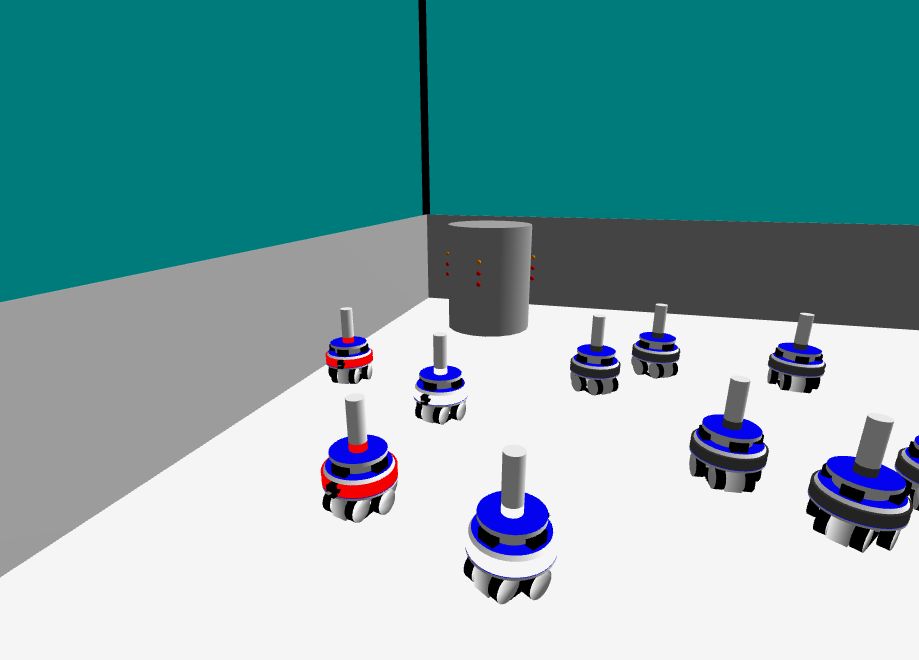}
    \includegraphics[width=0.25\columnwidth]{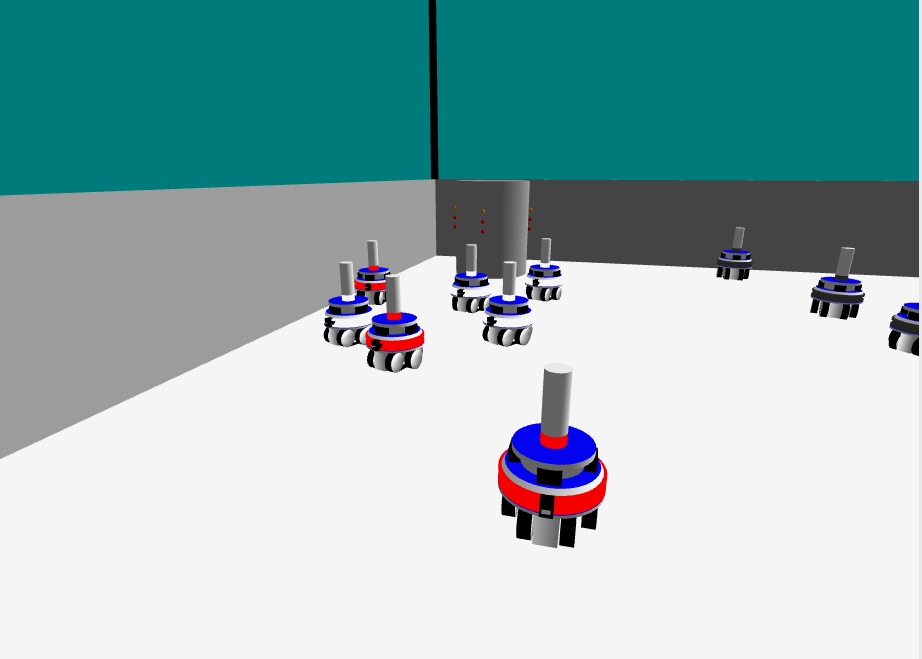}
    \caption{Subgoal formation process from subgoal}
    \label{fig:subgoalformation2}
\end{figure}

A subgoal robot can also serve as a robot beacon, allowing other robots to explore and detect it as their goal. This robot beacon, referred to as a subgoal-member, communicates with other robots by emitting a color signal through its LED ring. When a robot becomes a subgoal, it emits the color red. This distributed process results in the formation of one or more subgoal robots. The process continues until the robots reach the starting point, ultimately forming a complete path with intermediate subgoals from the goal to the starting point. This process is depicted in Figures \ref{fig:subgoalformation1} and \ref{fig:subgoalformation2}.

\begin{figure}
    \centering
    \includegraphics[width=0.5\columnwidth]{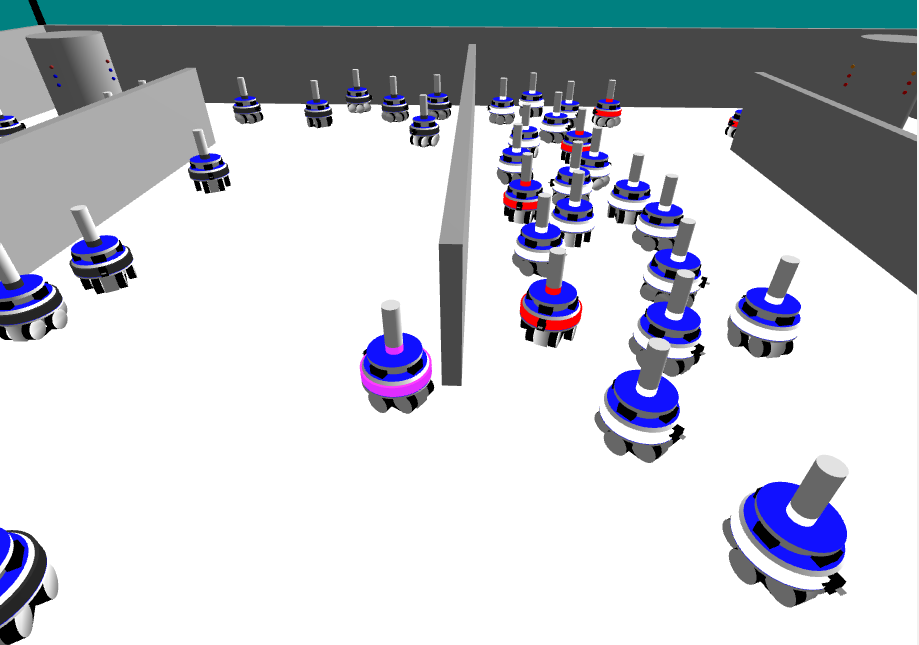}
    \caption{Recovery behavior/ hidden location identifier}
    \label{fig:recoveryrobot}
\end{figure}

One unique aspect of our research is the introduction of recovery robot behavior. Although robots can become subgoals within a range of 70, there may be cases where the goal/subgoal is not visible within that range. In such situations, the robot continues moving until it reaches the maximum visible range of 100. If the robot loses visibility of the goal/subgoal within its visibility range, it assumes that there is an obstacle between the robot and the goal/subgoal. Consequently, the robot switches to the recovery robot state, as illustrated in Figure \ref{fig:recoveryrobot}. Recovery robots inform other robots to avoid entering the invisibility area. If a robot detects a recovery robot within a range of 20, the recovery robot repulses it, ensuring that the robot avoids entering the blind spot. In certain cases, the robot may enter the repulsion range, triggering a state change and eventually becoming a subgoal.
 
\begin{figure}
    \centering
    \includegraphics[width=0.5\columnwidth]{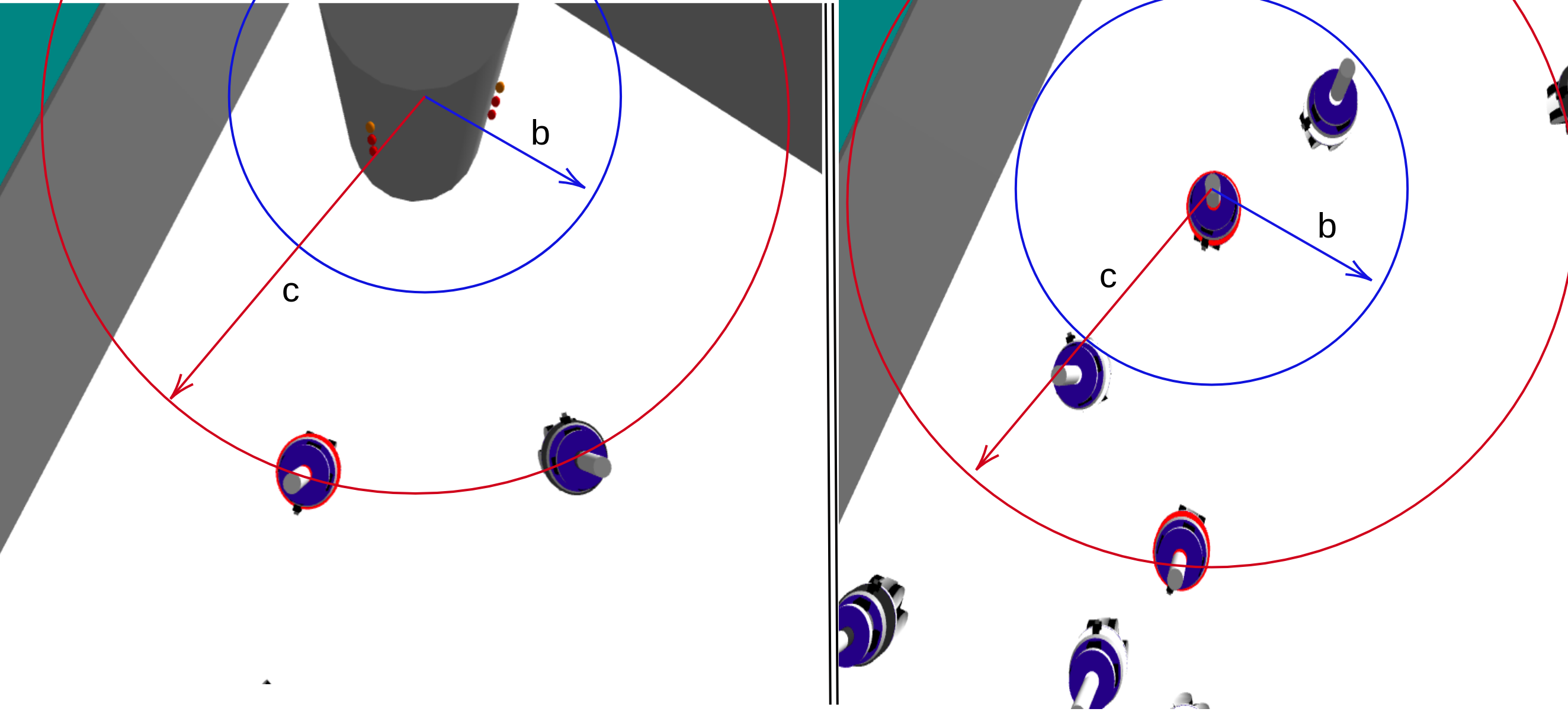}
    \caption{Descriptions of sub goal based path formation}
    \label{fig:parameter2}
\end{figure}

\subsection{Path formation stratagies}
\begin{figure}
    \centering
    \includegraphics[width=0.6\columnwidth]{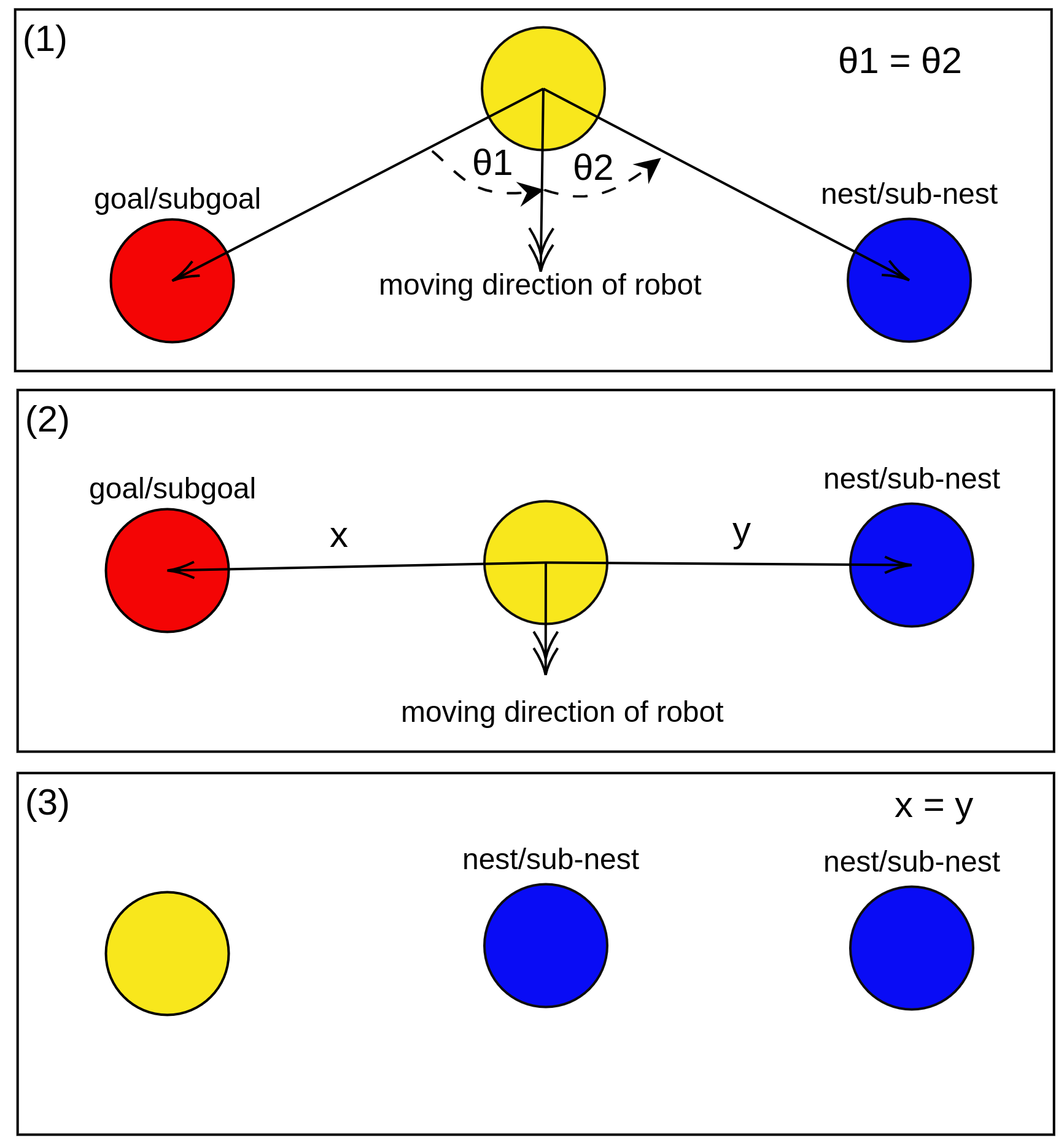}
    \caption{Heuristic Optimization 1}
    \label{fig:optimization1}
\end{figure}

\begin{figure}
    \centering
    \includegraphics[width=0.6\columnwidth]{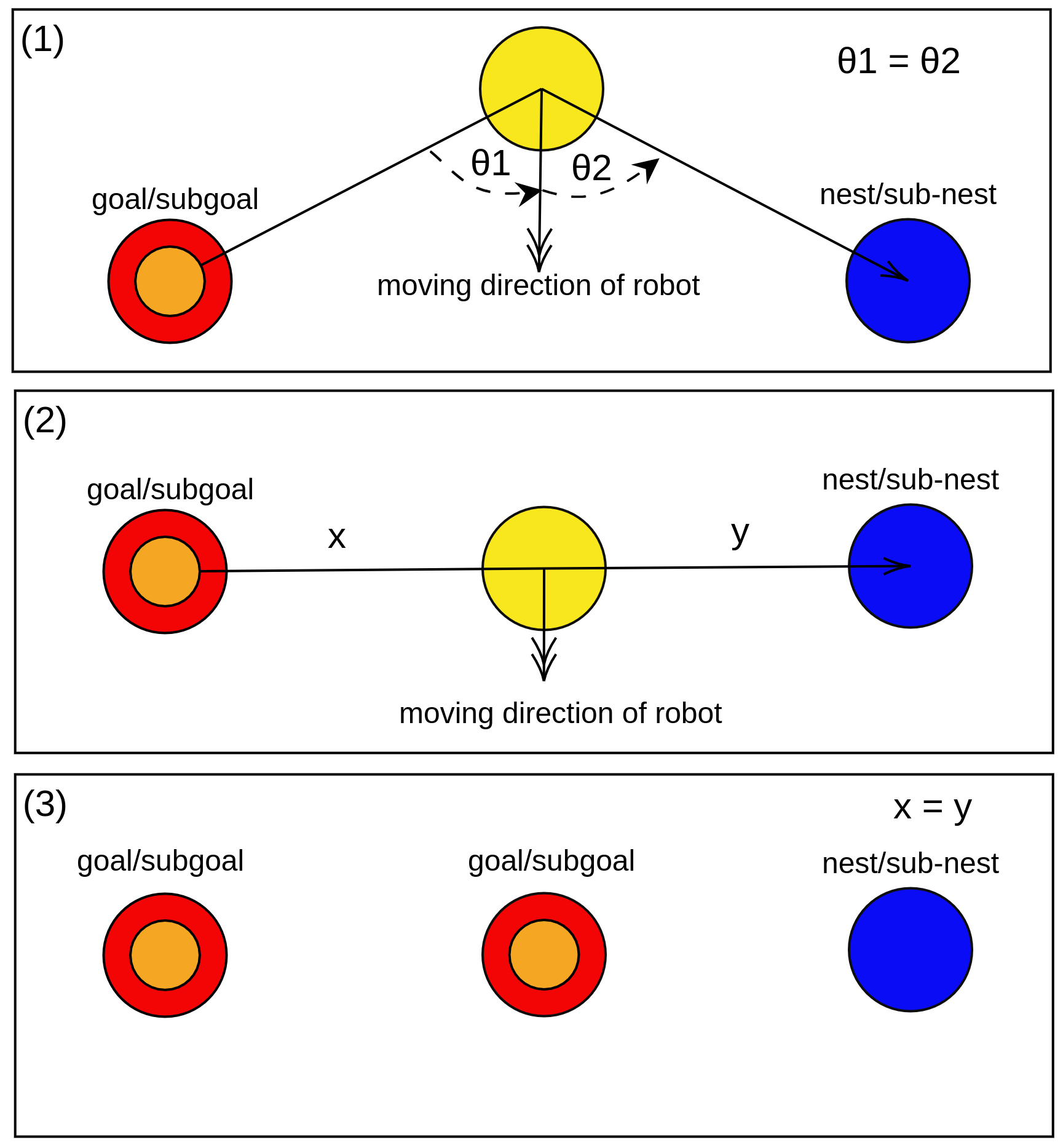}
    \caption{Heuristic Optimization 2}
    \label{fig:optimization2}
\end{figure}

The path formation strategies in our approach involve two heuristic optimization processes: one from the starting point to the goal, and another from the goal to the starting point. The optimization process begins when a subgoal robot detects the nest (represented by the color blue). The first subgoal robot from the nest initiates the first alignment process with the second subgoal robot. Once this process is successfully completed, the first subgoal robot starts emitting the color blue, indicating that it is acting as a sub-nest. This process continues with subsequent subgoal robots until the last subgoal robot is reached. Four parameters are utilized in the first optimization strategy, as depicted in Figure \ref{fig:optimization1}: $\theta_1$ represents the goal/subgoal angle, $\theta_2$ represents the nest/sub-nest angle, while $x$ and $y$ denote the distances between the goal/subgoal and the processing robot, and between the nest/sub-nest and the processing robot, respectively. The first optimization process involves adjusting the robot's position to minimize the error angle.
Once the last subgoal robot completes the first alignment process, it proceeds to perform the second alignment process from the goal to the nest. This second process continues until it reaches the first subgoal robot from the starting position, as illustrated in Figure \ref{fig:optimization2}. In cases where the alignment robot loses visibility of the subgoal or sub-nest within a certain visibility range during the first or second alignment process, it transitions to the recovery robot state. The role of the recovery robot is to inform other robots to avoid entering the invisibility area while they are in the subgoal formation process.

Our control system has been designed to ensure the desired behavior described above is achieved.
\subsection{Task allocation model}
\begin{figure}[h!]
    \centering
    \includegraphics[width=0.9\columnwidth]{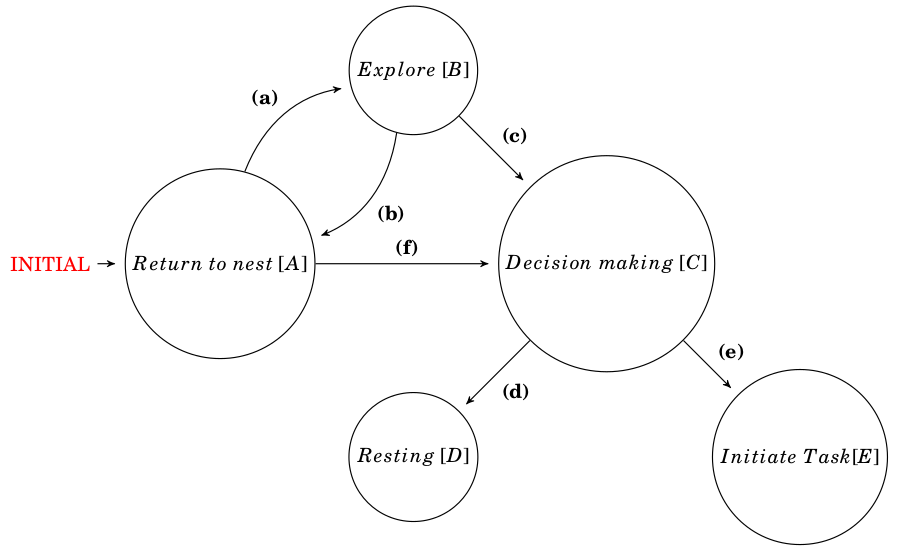}
    \caption{Finite state machine for proposed task allocation method}
    \label{fig:fsm}
\end{figure}

Our task allocation model is designed to ensure effective path formation while utilizing resources efficiently and aiming to create the shortest path. Instead of deploying all robots for the path formation task, which could lead to decreased performance due to traffic congestion, our method allocates only the required number of robots for this task. The remaining robots are assigned to the resting task. To determine the number of robots needed to form the path from the start to the end point, we employ a finite state machine (FSM) as depicted in Figure \ref{fig:fsm}.
\begin{table}[h!]
\caption{States description of proposed task allocation model}
    \renewcommand{\arraystretch}{1.5}
	\begin{tabular}{|r|p{2in}|} 
	\hline
	States&Description\\
	\hline
		Return to Nest&Returning to nest is the state to bring robots to	starting places through this we can keep the robots from lost \\
		Exploring& Searching for the goal from the starting point(nest).searching will happen opposite potential field from the starting point.Robots have max step sizes if robots fail to find the goal point then the robots change the state to returning to nest \\
		Decision making&When a robot find a goal or robots find goal finding robot go to this state. Here robots decide whether to go path formation or resting based on local communication protocols.\\
		Initiate task&In this state robots go to subgoal based path formation task \\
		Resting& If allocating Robots to path formation task reach it's optimal level other robots going to resting state \\
	\hline
	\end{tabular}
	\label{tab:State Table}
\end{table}
\begin{table}[h!]
\caption{State Transition of proposed task allocation model}
    \centering
    \begin{tabular}{|r|p{2in}|}
	\hline
	Transitions&Description\\ 
	\hline
		a&Explore to find goal/target until reach minimum exploring time\\
		b& If target not found in that minimum exploring time return to nest and increase the minimum exploring time to explore more. \\
		c& Found target and return to nest for making decision about how many robots need to go for path formation for that minimum exploring time\\
		d&Decided to do the resting task\\
		e& Decided to go for path formation\\
		f&Get information in nest about target is already found\\
	\hline
	\end{tabular}
	\label{tab:Transitions}
\end{table}
Local interactions among robots are facilitated through light signal-based communication. During the exploration phase, robots can detect the goal within their range of vision. If a robot detects the goal, it changes its color to indicate that it has found the goal to other robots. If the goal is not found, the robot moves towards the starting point based on the potential field.
\begin{equation}
l=s*t 
\end{equation}
The length of the path is calculated by robots using the exploration time and their speed. Equation (1) represents the calculation of the path length (l) as the product of the exploration time (t) and the robot speed (s). This information is used by the robots to determine when their energy level has reached its optimum, indicating that they should return to the nest.
\begin{equation}
n_0=l/v
\end{equation}
Based on the calculated path length (l) and the robot's visual range (v), the number of robots needed to form the path ($n_0$) is determined using Equation (2). However, $n_0$ is not directly applied to perform the path formation task due to the complexity of the environment and other factors involved in subgoal-based path formation. In such cases, some recovery robots are required for the subgoal-based path formation task, while other robots assume the role of subgoals without participating in the path. To account for these behaviors, a fixed factor based on the complexity of the environment, denoted as $\delta$, is added in Equation (3).
\begin{equation}
n=l/v+\delta
\end{equation}
The value of $\delta$ depends on the type of environment, allowing for flexibility in adapting the task allocation model to different scenarios.
\subsection{Local Communication Protocol}
Once the goal is found and a robot becomes the goal founder, the other robots enter the decision-making state through light signal-based interactions. They all proceed to the start point and initiate the process of determining which robots will be assigned to the path formation task and which ones will perform the resting task. This is where the local communication protocol comes into play.
\begin{figure}
    \centering
    \includegraphics[width=1\columnwidth]{ 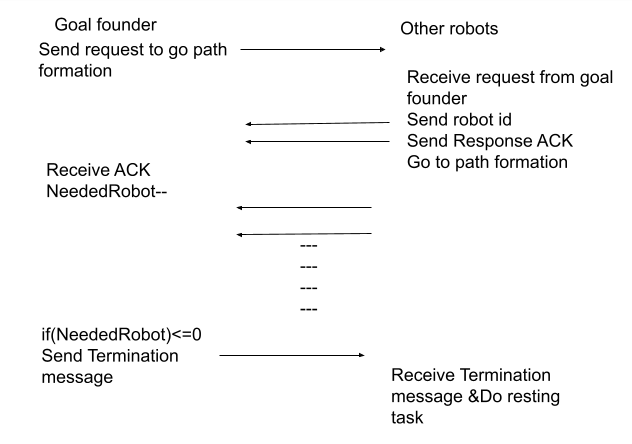}
    \caption{Communication protocol}
    \label{fig:protocol}
\end{figure}
The task allocation process utilizes a local communication protocol that includes both broadcast and unicast signals. As illustrated in Figure \ref{fig:protocol}, the goal founder robot broadcasts a signal to instruct other robots to engage in the path formation task. Upon receiving this request, the other robots respond with unicast signals containing their robot IDs, indicating their readiness to participate in the path formation task. The goal founder robot acknowledges the receipt of these signals and decrements the count of required robots. Once the desired number of robots has been allocated, the remaining robots transition to the resting task, and the communication process is terminated.
\begin{figure}
    \centering
    \includegraphics[width=0.5\columnwidth]{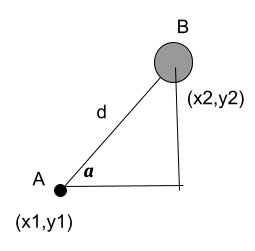}
    \caption{Resting robot moving to deployment point}
    \label{fig:movetorest}
\end{figure}

\begin{equation}
    d=\sqrt{(x_1-x_2)^2+(y_1-y_2)^2}
\end{equation}
  \begin{equation}
\alpha={sin}^{-1}{\left(\frac{|y1-y2|}{d}\right)}
\end{equation}
To prevent collisions between the path formation robots and the resting robots, the resting robots remain at the deployment point. To achieve this, a potential field is created towards the initial deployment point using euclidean distance, as described in Figure \ref{fig:movetorest}. The initial deployment point (x1, y1) and the current position (x2, y2) are obtained from the positioning sensors. The euclidean distance between the start and current positions is calculated using Equation (4). Furthermore, the angle $\alpha$ between the start position and the current position with respect to the y-axis is determined using Equation (5). With this angle $\alpha$ and the robot's heading angle, a potential field is generated to guide the robot towards the initial deployment point. Consequently, the robots utilize this potential field to move towards their respective initial points for resting, where they ultimately come to a rest.
\section{Evaluation and Comparisons}
\begin{figure}
    \centering
    \includegraphics[width=0.5\columnwidth]{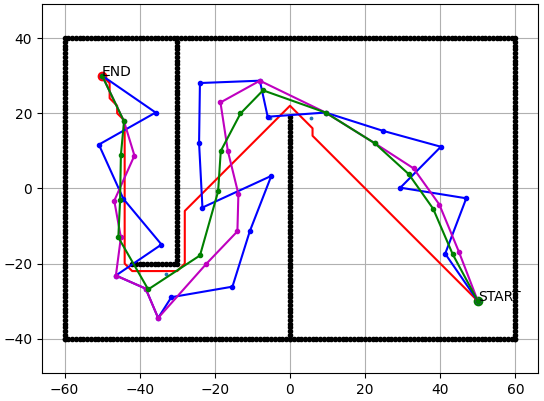}
    
    \caption{Comparison of optimization processes, subgoal  path with A*}
    \label{fig:comparison}
\end{figure}

The evaluation and comparisons of the testing model were conducted in eight different environments, including three open environments, three obstacle environments, and two complex environments. Each environment was tested with varying robot counts ranging from 60 to 100 robots. Figure \ref{fig:comparison} provides a visual comparison of the subgoal state path (blue), optimization 1 path (purple), optimization 2 path (green), and the A* algorithm path (red) across the 40 test cases.
\begin{figure}
    \centering
    \includegraphics[width=0.4\columnwidth]{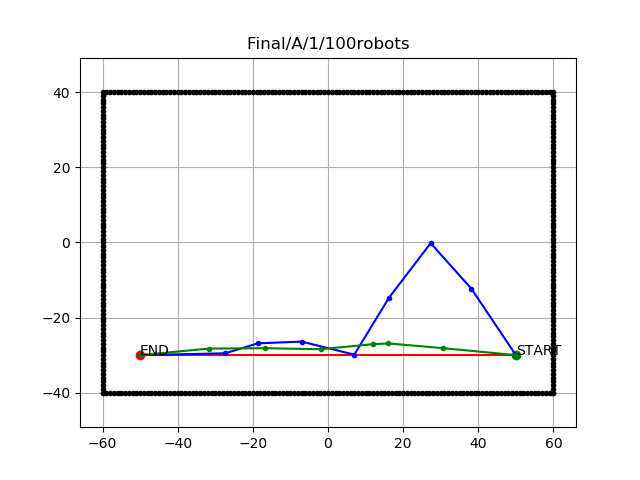}
    \includegraphics[width=0.4\columnwidth]{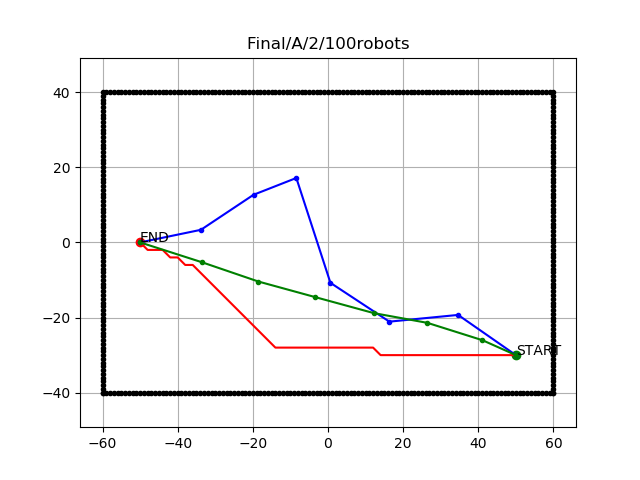}
    \includegraphics[width=0.4\columnwidth]{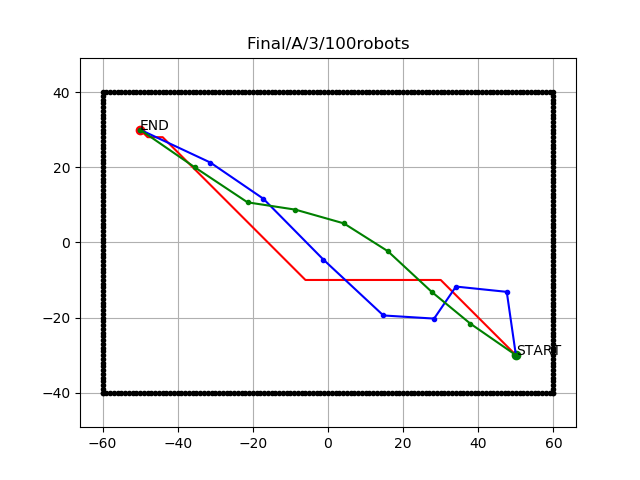}
    \caption{100 robots in open environment}
    \label{fig:1}
\end{figure}

\begin{figure}
    \centering
    \includegraphics[width=0.4\columnwidth]{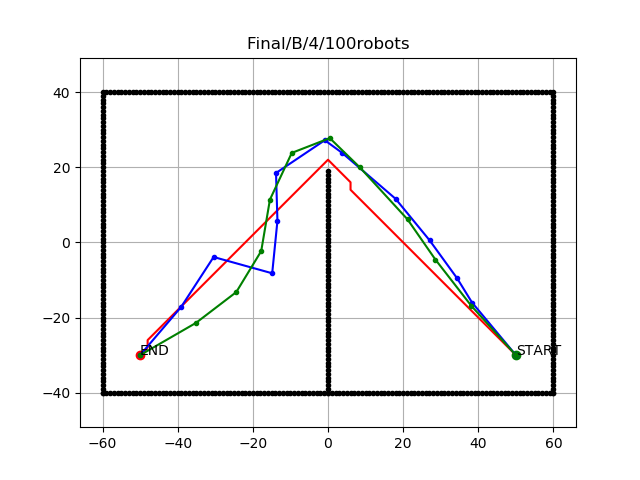}
    \includegraphics[width=0.4\columnwidth]{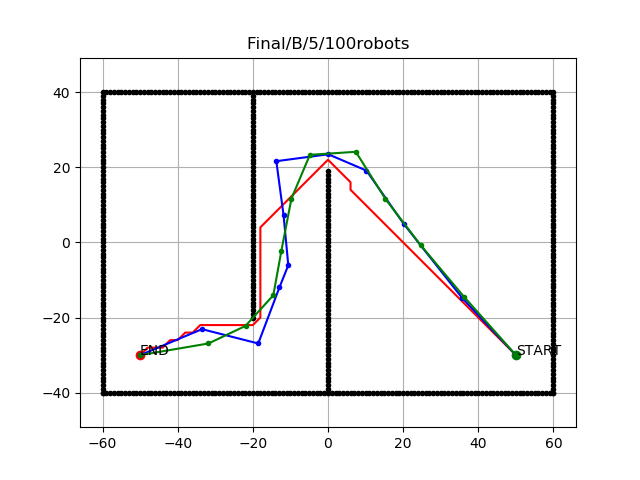}
    \includegraphics[width=0.4\columnwidth]{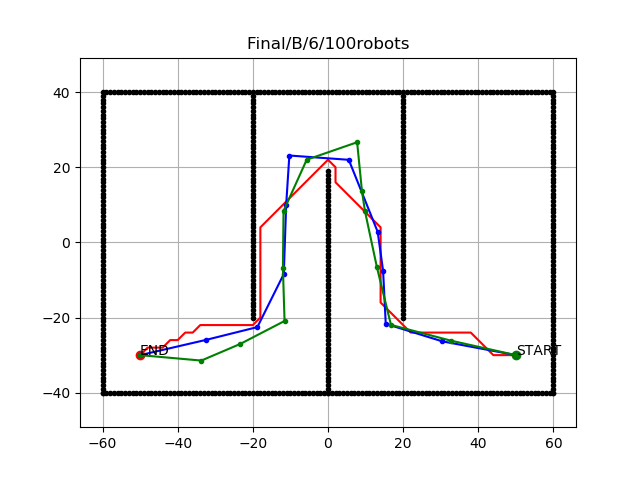}
    \caption{100 robots in obstacle environment}
    \label{fig:2}
\end{figure}
\begin{figure}
    \centering
    \includegraphics[width=0.4\columnwidth]{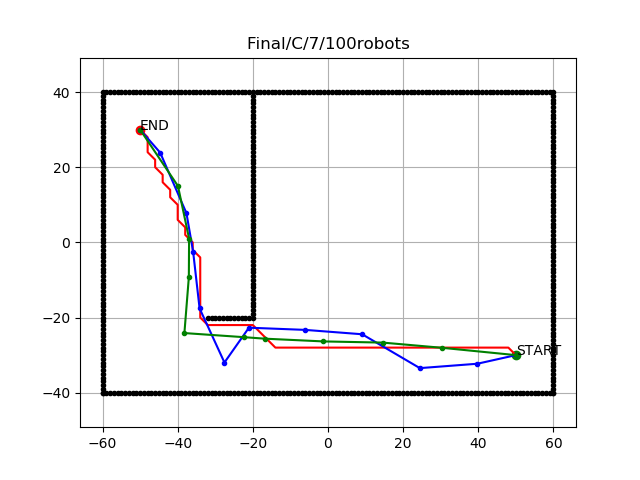}
    \includegraphics[width=0.4\columnwidth]{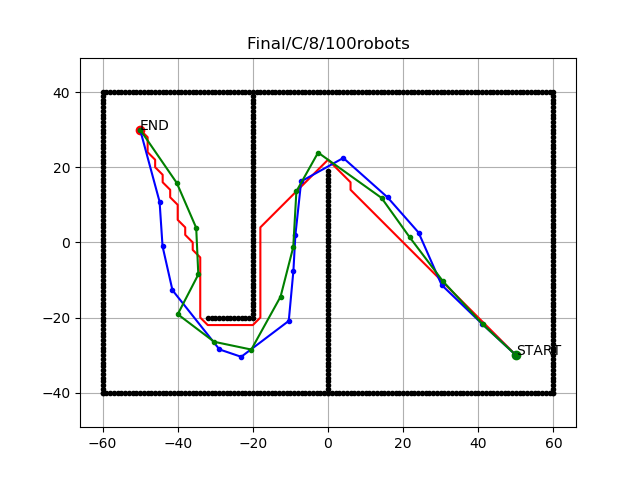}
    \caption{100 robots in complex obstacle environment}
    \label{fig:3}
\end{figure}
In Figures \ref{fig:1}, \ref{fig:2}, and \ref{fig:3}, the A* algorithm path is represented in red, the sub-goal formation final path is represented in blue, and the result obtained from improving the path using task allocation strategies is represented in green. The plotted dots in each graph indicate the locations of the robots as intermediate sub-goals. The success rate of sub-goal formation path formation was evaluated based on different environment types and robot sizes. The final path was compared with the sub-goal path, optimization 1 path, optimization 2 path, and the A* algorithm path.
\begin{table}[h!]
\caption{Test Results(average) for eight different environments}
\begin{tabular}{|p{0.4cm}|p{0.84cm}|c|c|c|c|p{1cm}|p{0cm}|} 
    \hline
    Env&\multicolumn{2}{|c|}{Time Taken}&\multicolumn{3}{|c|}{Path length}&Resource Reduction \\ 
    \hline
  {}& without&with&A-star&without&with&{} \\ 
  \hline
  1 & 4677 & 5047&100&131.4&100.3&77.1 \\ 
  \hline
  2& 7348 & 2389&112.8&147.5&104.4&75.4\\ 
  \hline
  3 & 5977 & 5026&132.9&149.5&118.6&75.4 \\ 
  \hline
  4 & 9754 & 8383&144.3&197.6&158.2&52.5 \\ 
  \hline
  5 & 10124 & 8164&158.9&150.3&146.2&49.2\\ 
  \hline
  6 & 14471 & 15280&172.5&201.0&183.1&47.1 \\ 
  \hline
  7 & 10464 & 7679&143.1&142.3&145.9&76.2\\ 
  \hline
  8 & 20123 & 19770&191.2&211.4&209.2&42.9 \\ 
  \hline
\end{tabular}
\end{table}
The performance of the task allocation model was also tested against the A* algorithm and the path before the implementation of the task allocation strategies. Table 5 presents the time taken to form the path in Argos default step size, the path length in Argos default unit, and the percentage of resource reduction achieved in the eight different environment types. The "Without" column refers to the path formation model without improvement through task allocation, while the "With" column represents the path after incorporating the task allocation model.

In the evaluation, the model without task allocation demonstrated that 25\% of the paths formed using the sub-goal based path formation were shorter than those generated by the A* algorithm. The success rate of forming paths without task allocation across the 40 test cases was 80\%.

Regarding the task allocation model, resource efficiency was calculated based on the deployed and allocated robot counts. On average, across the 40 different test cases, the model achieved a 61.93\% reduction in resource utilization. Path efficiency was evaluated based on the path length, with 40\% of the test cases forming paths shorter than those generated by the A* algorithm. In all 40 test cases, the path formed using the model with task allocation was shorter than the path formed without task allocation. Furthermore, 87.5\% of the cases with task allocation formed paths more quickly than those without task allocation.
\section{Conclusions and future work}
In this work, we have tackled the challenge of collective exploration and navigation using a swarm of robots. Our approach involved the development of a behavior-based controller inspired by foraging behavior, which relied solely on local information. We implemented and analyzed three different control strategies: the subgoal strategy, the aligning strategy 1, and the aligning strategy 2.

The subgoal strategy served as the foundation, resulting in the formation of static subgoals. The aligning strategy 1 extended this approach by introducing adjustments to the position of subgoal members, aiming to achieve specific distances and angles with respect to their neighbors. This led to the alignment of subgoals from the start to the goal. The aligning strategy 2 further expanded on aligning strategy 1, incorporating recovery robots to maintain a certain distance from obstacles and walls. Our algorithm dynamically adapted the path in any type of environment, ensuring robustness.

Furthermore, we proposed task allocation mechanisms for swarm subgoal-based path formation. Through light signal-based interactions, the robots initially explored the environment to locate a goal. Subsequently, communication protocols were employed during the decision-making phase to effectively allocate tasks. The task allocation model successfully utilized robot resources by allocating only the necessary number of robots for path formation tasks. This approach reduced resource requirements and deployment costs, allowing for the parallel utilization of excess robot resources in other tasks.

Comparisons with the A* algorithm and the model without task allocation demonstrated that our proposed model consistently formed the shortest paths. Future work could involve the implementation of more advanced communication protocols and testing the model with real robots in real-world environments. This would provide further validation and insights into the practical applicability of our approach.

\vspace{12pt}


\begin{thebibliography}{00}


\bibitem{b1} Jianing Chen, Melvin Gauci, Wei Li, Andreas Kolling and Roderich Grob, "Occlusion-Based Cooperative Transport with a Swarm of Miniature Mobile Robots",IEEE Transactions on Robotics, 2015.
\bibitem{b2} Gross Roderich and Dorigo Marco, "Towards group transport by swarms of robots",Bio-Inspired Computation, Vol. 1, Nos. 1/2, 2009.
\bibitem{b3} S. Nouyan, A. Campo, and M. Dorigo ,"Path formation in a robot swarm Self-organized strategies to find your way home",IRIDIA, CoDE, Universite Libre de Bruxelles, Brussels, Belgium, 2004.
\bibitem{b4} J Werfel ,"Collective construction with robot swarms,Morphogenetic Engineering", Springer,2012.
\bibitem{b5} K. Lerman. and A. Galstyan," Two paradigms for the design of artificial collectives", In Proceeding of the First Annual workshop on Collectives and Design of Complex Systems,
NASA-Ames, CA,2004.
\bibitem{b6} L. Panait and S. Luke," Cooperative multi-agent learning: the state of the art",  Autonomous Agents and Multi-Agent Systems, 11(3):387-434, 2005.
\bibitem{b7} Trullier, S. Wiener, A. Berthoz, and J. Meyer," Biologically-based artificial navigation systems: Review and prospects. Progress in Neurobiology", 51:483-544, 1997.
\bibitem{b8} Reynolds, Craig W.Flocks, herds and schools," A distributed behavioral model", ACM SIGGRAPH Computer Graphics, 1987.
\bibitem{b9} M. Bonani, V.Longchamp and S.Megnenat, "The MarXbot, a Miniature Mobile Robot Opening new Perspectives for the
Collective robotic Research," Int. Conf. Intell. Robot. Syst. IROS, Taiwan, pp. 4187-4193, 2010.
\bibitem{b10} A. Jevtic, A.Gutierrez, D. Andina and M. Jamshidi ,"Distributed Bees Algorithm for Task Allocation in Swarm of Robots",IEEE system Journal,June 2012.
\bibitem{b11} A. Brutschy, G. Pini, C. Pinciroli, M. Birattari, and M. Dorigo, "Self-organized Task Allocation to Sequentially Interdependent Tasks in Swarm Robotics", IRIDIA Technical Report Series,May 2012.
\bibitem{b12}  P. Giovanni , A. Brutschy, M. Frison,
A. Roli, M. Dorigo, and M. Birattari, "Task partitioning in swarms of robots: An adaptive method for strategy selection", IRIDIA Technical Report Series,May 2011.
\bibitem{b13} F. Ducatelle, A. Forster, G.A. Dicaro and L.M. Gambardella, " New task allocation methods for robotic swarms", Proceedings of the 9th IEEE/RAS Conference on Autonomous Robot Systems and Competitions, May 2009.
\bibitem{b14} Yang Y ongming, Chen Xihui, Li Qingjun Tian Yantao, "Swarm Robots Task Allocation Based on Local Communication", 2010 International Conference on Computer, Mechatronics, Control and Electronic Engineering (CMCE).
\bibitem{b15} J. Chen, M. Gauci, Wei Li, A. Kolling , " Occlusion-Based Cooperative Transport with a Swarm of Miniature Mobile Robots", IEEE Transactions on robotics, vol. 31, April 2015.
\bibitem{b16} C. Pinciroli, V. Trianni,  R. Grady, G. Pini, A. Brutschy, M. Brambilla, N. Mathews , E. Ferrante, C. Gianni, F. Ducatelle, M. Birattari, L. Maria, M. Dorigo "ARGoS: a modular, parallel, multi-engine simulator
for multi-robot systems", Springer Science+Business Media New York 2012.
\bibitem{b17}  A.Campo, S. Nouyan, M. Birattari, G.Roderich, M. Dorigo. "Negotiation of goal direction for cooperative transport", IRIDIA – Technical Report Series, April 2006.
\bibitem{b18} Wonki Lee, Neil Vaughan and Daeeun Kim "Task Allocation into a Foraging Task with a Series of Subtasks in Swarm
 Robotic System",IEEE TRANSACTIONS and JOURNALS, June 2020.
\bibitem{b19}  Bao Pang, Yong Song , Chengjin Zhang, Hongling Wang, and Runtao Yang, "Autonomous Task Allocation in a Swarm of Foraging Robots: An Approach Based on Response Threshold Sigmoid Model", International Journal of Control, Automation and Systems 17(X) (2019).
\bibitem{b20} D.Jha,  "Algorithms for Task Allocation in Homogeneous Swarm of Robots",  IEEE/RSJ International Conference on Intelligent Robots and Systems (IROS), October 2018.
\bibitem{b21} J. Zhou, M. Dejun , F. Yang, D. Guanzhong, "Labor division for swarm robotic systems with arbitrary finite number of task types",Proceeding of the IEEE International Conference on Information and Automation Hailar, China, July 2014.
\bibitem{b22} S. Goss, S. Aron , L. Deneubourg, J. Pasteels, "Self orgaized shortcuts in the Argentine ant",Springer, Decemer 1989.
\bibitem{b23} D. Payton, M. Daily, R. Estkowski, M. Howard, C. Lee "Pheromone Robotics",Springer, November 2001.





\end{thebibliography}
\end{document}